# Axe the X in XAI: A Plea for Understandable AI

Andrés Páez
Universidad de los Andes
*apaez@uniandes.edu.co*

**ABSTRACT**

In a recent paper, Erasmus et al. (2021) defend the idea that the ambiguity of the term "explanation" in explainable AI (XAI) can be solved by adopting any of four different extant accounts of explanation in the philosophy of science: the Deductive Nomological, Inductive Statistical, Causal Mechanical, and New Mechanist models. In this chapter, I show that the authors' claim that these accounts can be applied to deep neural networks as they would to any natural phenomenon is mistaken. I also provide a more general argument as to why the notion of explainability as it is currently used in the XAI literature bears little resemblance to the traditional concept of scientific explanation. It would be more fruitful to use the label "understandable AI" to avoid the confusion that surrounds the goal and purposes of XAI. In the second half of the chapter, I argue for a pragmatic conception of understanding that is better suited to play the central role attributed to explanation in XAI. Following Kuorikoski & Ylikoski (2015), the conditions of satisfaction for understanding an ML system are fleshed out in terms of an agent's success in using the system, in drawing correct inferences from it.

## 1. Introduction

The concept of explanation has a long and variegated history in the philosophy of science. Starting with Hempel and Oppenheimer's (1948) logic-based approach, at least a dozen different analyses of the concept have been proposed.[1] Given this rich theoretical repository, some philosophers have argued that the solution to the problem of specifying a meaning for "explanation" in the context of artificial intelligence (AI) is to adapt an extant account of scientific explanation to machine learning (ML) in general, and to deep neural networks (DNNs) in particular. Erasmus et al. (2021) offer the most developed account of this strategy. They examine four different accounts of explanation in the philosophy of science: the Deductive Nomological, Inductive Statistical, Causal Mechanical, and New Mechanist models. Their claim is that any of them is applicable to DNNs as it would to any scientific phenomenon. This claim derives from a more general principle that they call "the indefeasibility thesis" about explanation (p. 840). The thesis states that explanations are invariant with respect to the complexity of both the explanans and the explanandum. There is no threshold of complexity beyond which a phenomenon becomes unexplainable. Therefore, despite their complexity, DNNs are scientifically explainable.

---

[1] See Woodward and Ross (2021) for a survey of the earliest models of explanation, roughly until 1990, and Ross and Woodward (2023) for more recent causal approaches.

In this chapter, I argue that the thesis that opaque ML systems are scientifically explainable is either trivial or false, and that it misrepresents the goals of explainable AI (XAI). It is trivial if an explanation is simply understood as the set of causes, entities or states that physically or computationally produce a prediction; not the linguistic or mathematical *description* of the known elements in the set, but the elements themselves, known or unknown. It is false if the claim is that it is always possible to offer an "explanatory text," a *truthful* description of the source of the prediction, thereby satisfying the factivity condition on scientific explanations (Páez 2009, 2019). Most of the first half of this chapter will be devoted to justifying the second claim. Now, if there are explanatory gaps in machine learning, and more specifically, if the predictions of DNNs cannot be scientifically explained, then the goal of *explainable* AI thus formulated will be unattainable. We should not insist on using a concept that cannot perform its desired function. Instead, I will advocate for the use of labels such as "interpretable machine learning" (Watson & Floridi, 2021) or "understandable AI" to avoid the confusion that surrounds the goal and purposes of XAI methods. In the second part of the chapter, I argue that understanding is a success concept that is better suited to play the role often attributed to explanation in XAI. Following Kuorikoski & Ylikoski (2015), the conditions of satisfaction for understanding an ML system are fleshed out in terms of an agent's success in using the system, in drawing correct inferences from it.

## 2. Clarifying the Explanandum

Before we can tackle the question of explanation in the context of machine learning, it is essential to clarify what it is that we are trying to explain, that is, to clarify the explanandum. There are two possible explananda in ML. From the point of view of local *post-hoc* interpretability methods, the explanandum is some specific prediction churned out by a trained model. These methods usually try to pinpoint input features that make a difference in the prediction. Local interpretations include counterfactual probes (Wachter et al., 2018; Mothilal et al., 2020), and different types of perturbation-based methods such as LIME (Ribeiro et al., 2016;), Grad-CAM (Selvaraju et al., 2017; Ancona et al., 2019), SHAP (Lundberg & Lee, 2017), TCAV (Kim et al, 2018), among many others (see Ivanovs et al., 2021 for a survey).

From the point of view of *global* interpretation methods, the explanandum is the entire model, more specifically, the non-linear function performed by the trained model. Global interpretations generally take the form of surrogate models. The most widely used classes of surrogate models are linear or gradient-based approximations, decision rules, and decision trees (Frosst & Hinton, 2017; Wu et al., 2018). Some surrogate models are generated through knowledge distillation techniques (Bastani, 2019; Jung et al., 2017; Kim et al., 2022; Tan et al, 2018). Given a model *f*, the goal is to generate an interpretable model M such that $M(x) \approx f(x)$. A different approach is to use example-based methods. The idea is to select subsets of the dataset to explain the behavior of ML models or to make explicit the underlying data distribution. This approach only works when the data is structured and can be represented in a human understandable way. They include similar examples or factuals (Schoenborn et al., 2021), influential instances (Koh et al., 2017, 2019) and prototypes (Kim et al., 2016), among others. These methods yield results that are more easily understood.

Most theories of explanation in the philosophy of science have taken individual facts or particulars, such as things or events, as their explanandum. Thus, the explanandum of a scientific explanation is a true statement that expresses a fact or that states that the things or events to be explained exist or occur. In that respect, they resemble *post hoc* local interpretability methods. Kitcher's (1989) unificationist model of explanation uses a different approach. Scientific explanation is a matter of providing a unified account of a range of different phenomena. Erasmus and Brunet (2022) discuss

Kitcher's unificationist model in the context of ML, but their argument is parasitic on their defense of the applicability of Deductive-Nomological explanations to DNNs. I will therefore skip the specifics of that discussion to focus on the four theories of explanation that seem more readily applicable in the context of ML.

Notice that if we restrict the discussion to these four models of scientific explanation, whose explananda are always individual facts or events, global explanations will be excluded because the target of any global interpretation method in ML is the original model itself, not one of its predictions. As a result, ML models will be scientifically unexplainable. ML models are complex mathematical structures that admit of infinitely many input-output pairs; their abstract nature makes them akin to a law of nature, not to a singular fact or event. This means that the account provided by Erasmus et al. (2021) is incomplete at best because it cannot account for global explanations. The authors could try to amend this shortcoming by adding a model of explanation of laws to the mix, but this approach faces the problem that there is no agreed upon account of the laws of nature in the philosophical literature, let alone an account of the explanation of laws.[2]

It could be argued that although surrogate models do not explain singular predictions, they can be seen as explanatory *schemata*. Plugging in the missing values in a surrogate decision tree, for example, will provide an explanation of a singular prediction.[3] However, the price of taking this way out of the problem is that factivity will have to be sacrificed at the outset because surrogate models are always false, as Cynthia Rudin explains:

> Explanations must be wrong. They cannot have perfect fidelity with respect to the original model. If the explanation was completely faithful to what the original model computes, the explanation would equal the original model, and one would not need the original model in the first place, only the explanation (2019, p. 207).

---

[2] In "Studies in the logic of explanation," Hempel and Oppenheimer famously state that the explanation of laws "presents peculiar problems for which we can offer no solution at present" (1948, p. 165, fn. 33). These problems force them to limit their theory to the explanation of particular events. The situation is no better today than in 1948. I'm grateful to Juan M. Durán for reminding me of this telling passage.

[3] I am grateful to Stefan Buijsman for suggesting this possibility.

Since Erasmus and colleagues want to defend the factivity of explanation (see below), surrogate models, even when regarded as explanatory schemata, would not qualify as bona fide explanations.[4] The remaining question is whether their account is successful in the explanation of singular predictions.

## 3. Scientific explanation of ML systems?

In this section, I will examine whether the four theories of explanation discussed by Erasmus et al. (2021) can play the role attributed to them. I will focus mostly on the way they use Hempel's Deductive-Nomological (D-N) model because discussion of that model will bring out the main elements in play.[5] Once the stakes have been made clear, it will be much easier to dismiss the use of the other three models.

### 3.1 Deductive-Nomological Explanations

I will begin by examining the idea that Hempel's Deductive-Nomological model of scientific explanation can be used to explain the output of a DNN. Hempel's model can be represented by the well-known schema that deduces the explanandum from laws and initial conditions, which jointly constitute the explanans (1965, p. 336):

$$L_1, L_2, \ldots, L_n$$
$$\underline{C_1, C_2, \ldots, C_n}$$
$$E$$

Here $L_1, L_2, \ldots, L_n$ are general laws, $C_1, C_2, \ldots, C_n$ are sentences describing the particular facts involved, and $E$ is a sentence describing the explanandum phenomenon. As Hempel is well aware, general laws in science take many forms and even nowadays there is no agreed-upon definition of what a law of nature is. However, to have an explanatory character, Hempel argues, laws should not make reference to particulars. We must require them to be of "essentially general form" (p. 343). Finally, all the elements in the explanation must be true. This is generally known as the *factivity condition* on scientific explanation.

---

[4] Following Elgin (2017) and Potochnik (2017), in section 4, I argue that the factivity condition on scientific explanation can be replaced by a more pragmatic approach in which surrogate models only need to be "true enough" of their target models.
[5] For the purpose of this chapter, I will ignore the fact that Hempel's D-N and I-S models are nowadays considered museum pieces due to the well-known objections that have been raised against them. Salmon (1989) offers the best discussion of these objections and counterexamples.

Before we move on to examine whether D-N explanations can explain the output of a DNN, it is important to understand the theoretical claim that undergirds the very possibility of attempting such a move. As we saw in the Introduction, Erasmus and collaborators argue, essentially, that everything is explainable because the possibility of devising an explanation does not depend on the complexity of the phenomenon:

> The features that make something an explanation turn out to be invariant with respect to the complexity of both the explanans and the explanandum. … Adding complexity to the explanation of the phenomenon does not entail that the phenomenon is any less explainable. This is not a claim about the quality, superiority, or goodness of a given explanation. Our concern is whether increasing the complexity of a given explanation makes it no longer an explanation (2021, p. 840).

Let us examine how the thesis applies to D-N explanations. We will only consider an increase in the complexity of the explanans because in supervised ML the explanandum always remains simple, viz. a prediction or a classification.

How could more information be added to the explanans of a D-N explanation? Erasmus et al. argue that one could, for example, replace one of the law-like statements and substitute it for a set of *n* laws jointly entailing it (here E is a set of empirical conditions):

> This may make the explanans less manageable or more difficult to understand, but we still have a DN explanation. That is, if *x* follows deductively from L + E, then it also follows from L* = $\{L_1+L_2+...+L_n\}$ + E, where $L_{1...n}$ are (complexity-increasing) laws which, taken together, entail L. The expansion of the explanans does not make its laws any less law-like nor its conclusion any less deductively valid. Put another way, the explanandum is no less explainable, since the connection to the explanans, no matter [its] complexity, is still deduction (p. 841).

In the context of a DNN, this argument requires that we identify law-like statements of any degree of complexity to set up the explanans of a D-N explanation. Here we will focus on two points, the relata of the laws and the factivity condition. Regardless of the form of a lawlike statement in science, laws create (mostly mathematical) relationships between predicates. Such predicates might refer to natural classes such as the electric charge of a particle or the volume of a gas, and they are applied to the state of a system:

> The state of such a system at any given time is characterized by the values assumed at that time by certain quantitative characteristics of the system, the so-called variables of state; and the laws specified by such a theory for the changes of state are deterministic in the sense that, given the state of the system at any one time, they determine its state at any other, earlier or later, time (Hempel, 1965, p. 351).

The first question that arises, if one wants to adapt Hempel's D-N model to DNNs, regards the relata of the lawlike statements that will be deployed. Are they relationships between classes of inputs and outputs? Between features of the input space and outputs? Between the weights and biases in the neural net and an output? If we think of a *functional* explanation based solely on the inputs and outputs of the system, it is impossible to think of deterministic laws in DNNs. No class of inputs will inevitably produce the correct decision every time, and sometimes even repeating the sampling might produce a different decision (Kindermans et al., 2017; Molnar, 2020).[6] Something similar occurs if we think of a lawlike relation between types of features and decisions. Simply choosing some set of features from the input space to train the system will not guarantee a deterministic relation between the chosen features and the decision. For example, if a bank uses features such as income, age, credit history, number of family members, and so on to train a fairly accurate ML system, there is still no guarantee that those particular features will generate the same predictions when provided with the same inputs. The only viable alternative is to look at the model itself as composed of deterministic lawlike relationships. This seems to be what Erasmus et al. (2021) have in mind. The example they use is of a DNN trained for image classification. They explain how to apply a D-N

---

[6] To be sure, there are *functionally* deterministic models in ML, but they have many limitations that make them unsuitable in most real-world applications.

explanation of why the DNN classified an input image in one of the output classes in the following way:

> A DN explanation of how the [DNN] assesses an input image involves listing the weights attached to each and every node and the informational routes indicated by each and every edge at every convolution stage, and the weights of the fully connected network along with the assigned numerical values being fed into the input layer and the network architecture. Once we have that, we can list the values for the classifications the [DNN] learned in the training and testing phases of development, and see that its classification of the image is based on comparing the ranges of these classifications with the output value of the image. In doing so, we are explaining the explanandum—here, the [DNN] classifying of image *I* as classification *c*—using an explanans consisting of a law-like premises—in this case, how the weights of all relevant nodes and edges produced the output value, along with the law that an output is assigned to the most probable class—and additional information about *I*—which includes the set of input values assigned to *I*, and the output value *c* (p. 844).

In more simple terms, the D-N explanation of the explanandum value *c* consists of an exact description of the model, which would take the place of the lawlike sentences $L_1$, …, $L_n$ in Hempel's model, and a list of the features of the input image *I*, which would correspond to the sentences $C_1$, …, $C_n$ that describe the particular facts involved. Together, they explain why the model generated the output value *c*.

There are two problems with this idea. Firstly, the connection between the features of the input space and the parameters learned by the model cannot be made sense of in an opaque model. Secondly, since there is no direct access to the hidden layers of the model, the values of the weights and biases cannot be verified; therefore, there is no way to determine whether any given explanatory text is true. Let us examine each problem in turn.

Neural networks are designed to identify patterns and make predictions based on these patterns. Inputs pass through multiple layers of interconnected nodes, or "neurons," each of which transforms them in some way before passing them along to the next layer.

These transformations often involve nonlinear operations, such as ReLU (Rectified Linear Unit), sigmoid, or tanh functions. These nonlinear transformations, combined with the interactions between the neurons across different layers, allow the neural network to model complex, high-dimensional decision boundaries. Even if we could see all the weights and biases in the network (the parameters that the model learned during training), it is not clear how to interpret them in the context of the original input features. Each input of the network goes through a series of complex, intertwined transformations that make it difficult to understand how the inputs relate to the output. The nonlinear nature of the activation functions compounds this complexity. In a linear system, the effect of each input can be considered independently of the others: if you double an input, the output will also double. But in a nonlinear system, the effect of changing an input depends on the values of all the other inputs. This makes it much more difficult to understand how each input influences the output, and in any case the relationship cannot be expressed using a single lawlike statement. In a neural network, there is no clear sequence of "decisions" to follow —the output is the result of a complex, interconnected web of influences that do not lend themselves to an explanation.

The second problem is that there is no way to verify which parameters are being used in the hidden layers of the neural net. Deep models often have an extremely large number of optima of similar predictive accuracy. This is known as the *model identifiability problem*: "A model is said to be identifiable if a sufficiently large training set can rule out all but one setting of the model's parameters. Models with latent variables are often not identifiable because we can obtain equivalent models by exchanging latent variables with each other" (Goodfellow et al., 2016, p. 284). It is therefore impossible to verify which of many possible equivalent models is the one that generated the output in this case. Without identifying which model is being used, there is no way to create a true explanatory text of the model's prediction. Therefore, the factivity condition on explanation cannot be fulfilled. There *is*, of course, a true description of the model, but it lives in Popper's World 3, forever inaccessible to human subjects.

Independently of these two unsolvable problems, there is something amiss in saying that a ML model is a set of lawlike statements. It is true that the model is a function and most laws in science are mathematical functions, but that is as far as the resemblance goes. As mentioned above, Hempel insisted once and again that laws must have an essentially general form. They cannot be a conjunction of descriptions of particular facts:

> A statement which is logically equivalent to a finite conjunction of singular sentences, and which in this sense makes a claim concerning only a finite class of cases, does not qualify as a law and lacks the explanatory force of a nomological statement. Lawlike sentences, whether true or false, are not just conveniently telescoped summaries of finite sets of data concerning particular instances (1965, p. 377).

This statement alone is a rebuttal of Erasmus et al. There is nothing general in a trained ML model. A description of the model is exactly what Hempel describes: a conjunction of singular sentences about the weights and biases of the model. Its parameters obey the nature of the class of things in the training dataset and cannot be used for other ends. It is a custom-made predictive artifact that serves one and only one purpose. Furthermore, the features used to train the model do not necessarily correspond to natural kinds or to properties that belong to anything outside of the input space. Its limited scope in many cases is another reason why it lacks the generality of laws. It all depends on the data used. In the case of a model like AlphaFold, the features most probably correspond to natural kinds, but in most cases, the features can only be defined within a very specific context of use. ML just seems the wrong context to speak of laws, at least in the Hempelian sense, and a fortiori of covering law models of scientific explanation.

**3.2 Inductive-Statistical Explanations**

The second type of explanation discussed by Erasmus et al. (2021) is Hempel's inductive-statistical model (I-S). Unlike the D-N model, the I-S model uses statistical laws instead of deterministic ones. It preserves, however, the argument form of D-N explanations, but naturally in inductive form. The general schema is:

$$\frac{\begin{array}{c} L_1, L_2, \ldots, L_n \\ C_1, C_2, \ldots, C_n \end{array}}{E} \quad [r]$$

$L_1$, $L_2$, …, $L_n$ are general laws, but of statistical nature. They have the form $p(P|Q) = r$ and they must also lack any reference to particular instances. As before, $C_1$, $C_2$, …, $C_n$ are sentences describing the particular facts involved, and $E$ is a sentence describing the explanandum phenomenon. The explanans provides inductive support to the truth of the explanandum-statement. The strength of the support is indicated by the number in square brackets. According to Hempel, $r$ must be very high for the explanandum to be expected in light of the premises. As with D-N explanations, both the statistical laws and the empirical conditions must be true.

An I-S explanation of a prediction of a DNN model cannot be based on the model's parameters, as in the case of D-N explanations, because the trained model does not behave stochastically. The only options are to establish a statistical correlation between the input and output classes, or between the features of the input space and the output classes. Describing the first correlation in terms of a statistical law with a very high $r$ is the same as saying that the model is reliable, a claim that has no explanatory value regarding how a particular prediction was reached. For example, saying that the probability that a reliable cat-identifying model will identify a cat when presented with an image of a cat is very high adds nothing to our previous knowledge of the model's reliability. To be sure, the model's reliability leads to the expectation that the prediction is correct, but if most predictions were not correct, the model would not have been deployed in the first place. Therefore, the expectability of the output is the result of a correct validation procedure using the test set, and not of a *post hoc* inductive argument that offers no *independent* reason to expect the output.

Essentially, the same argument applies in the second case, viz. when the statistical lawlike correlation is established between a set of features and the output classes. The particular features selected in the training stage of a reliable DNN model are strongly correlated with its true predictions, but only because the model has been correctly validated. This option runs into an additional problem, that of identifying the features that are strongly correlated with the correct output. In a computer vision task, the features are largely detected by the learning algorithm and not preselected, as in the previous example of the bank loan model. Since the statistical law must be true, one must be able to identify those features. Local *post hoc* interpretability methods like LIME pinpoint the features responsible for an output for any given image, but the results cannot be generalized to other similar images, and they even differ when the sampling process is repeated. Without identifying such features, no true lawlike description of the correlation can be established. Simply saying that there is an epistemically inaccessible set of features responsible for raising the probability of a correct outcome is trivially saying that the process is not random (or magical). Finally, Hempel's point about laws not being a conjunction of singular sentences about specific facts, explained in the previous section, is just as valid in both of these cases.

So, how do Erasmus et al. (2021) apply I-S explanations to DNNs? First, they argue that the explanation will not be of individual predictions, but rather of the accuracy rate of the model. This runs counter to Hempel's original intention of explaining the high predictability of individual facts; their version of I-S explanations have an entirely different explanandum. An I-S explanation of the output class, in their terms, amounts to "details about the training process as statistical laws and the nature of the training data used as empirical information. If these, taken together, inductively entail a probability that the ANN's outputs are accurate, then we have successfully explained the accuracy of the ANN's outputs" (pp. 844-845). It is not clear at all how details about the training process can become statistical laws. No further details are provided. If what they have in mind are correlations between features and true predictions, their strategy has already been examined and refuted in the previous paragraphs. In any case, the authors have given up the game of using Hempel's original I-S model, so it is difficult to see what logical structure such an explanation would have. Presumably, it would still be a covering-law model because the use of statistical laws is retained.

**3.3 Mechanical Explanations and the New Mechanistic View**

The last two models of scientific explanation I will discuss are causal in nature. Talk about causation in ML can be seen from two different perspectives. On the one hand, one might want to understand the cause of a specific prediction in a pre-trained model, regardless of whether there is any real causal connection in the world between the identified features and the prediction. On the other hand, one might want to use ML to look for some of the causes of a real-world phenomenon, in which case the features picked up by the ML model must reflect the causal structure of the world. In the present discussion, we will only be concerned with the first use of causal connections.

In *Scientific explanation and the causal structure of the world*, Salmon (1984) proposes a causal mechanical model of explanation based on the idea of a causal *process*. A causal process is characterized by its ability to transmit a mark or the ability to transmit its own structure, in a spatio-temporally continuous way. The local transfer of energy and momentum, for example, constitutes a causal process, while the successive positions occupied by a shadow cast by a moving object do not. Causal *interactions* occur when one causal process spatio-temporally intersects another and produces a modification in its structure following conservation laws. A typical example is a collision of two particles. The sum of all causal processes and interactions forms the causal structure of the world mentioned in the title of the book. To explain a particular occurrence of a phenomenon is to show how it fits into the causal structure of the world. Salmon's original idea was later refined by Dowe (2000) in the conserved quantity model of causation.

How would Salmon's causal mechanical (CM) model be used to explain a prediction of an artificial neural network (ANN)? Erasmus et al. (2021) explain:

> For CM explanations of ANNs, we would cite the causal processes and causal interactions involved. This would entail describing the ANN using oft-used terms of art, drawing on biological analogies. ANNs are constituted by connected nodes, sometimes referred to as artificial neurons that, like biological neurons, receive and send information, or signals, via connections resembling biological synapses, termed edges (p. 845).

Causal talk in this case is purely analogical. The model is not a causal structure because the entire process is mathematical in nature. The "marks" that are transmitted are not physical quantities but approximations of real numbers that are processed at each node using some non-linear function of their weighted sum.[7] These are not CM explanations at all because no genuine causal interaction is involved in the exchange of information. This is not to say that causal analogies are useless in understanding a natural or artificial process. As several authors have argued, analogies are a pathway towards understanding complex phenomena (Bartha, 2010; Hesse, 1966). But analogies are not truthful representations of their target phenomena, and therefore fail the factivity condition for scientific explanation.

---

[7] See Stefan Buijsman's contribution to this volume (chapter 6) for further discussion of the mathematical nature of ML models and the need to think of ML explanations in non-causal terms.

Now, DNNs are grounded on genuine causal processes, namely, the electronic interactions in the hardware circuitry involved in computation. But even if there were a feasible way to access and comprehend these physical interactions, that would only add a new layer of uncertainty and complexity to the explanation since there would be no way to establish correlations between the genuine causal processes that take place in the hardware and the mathematical operations that take place in the hidden lawyers of the DNN. Furthermore, Durán (2018) has argued that computational results are underdetermined by physical interactions in the computer because each run of the algorithm instantiates a different set of physical elements in the computer. Erasmus et al. (2021, p. 846) seem to acknowledge that hardware processes must be treated as elementary, focusing instead on the DNN's architecture.

A similar analysis applies to the New Mechanistic view defended by Machamer et al. (2000). According to the New Mechanistic view, providing an explanation involves showing how some phenomena regularly arise from a collection of entities and activities. Both entities and activities are individuated through their spatiotemporally located physical properties, and causal processes are an essential component of any mechanism. Their approach is not far from Salmon's mechanical model: "Our emphasis on mechanisms is compatible, in some ways, with Salmon's mechanical philosophy, since mechanisms lie at the heart of the mechanical philosophy" (p. 7). Functional descriptions of a phenomenon, on the other hand, require an understanding of the underlying mechanism, since "functions are the roles played by entities and activities in a mechanism … [they] should be understood in terms of the activities by virtue of which entities contribute to the workings of a mechanism" (p. 6). Therefore, functional analyses have to be causally grounded.

Applying the New Mechanistic view to ML models yields descriptions at three different levels of abstraction: as a description of the functional correlation between inputs and outputs, as a "mechanical" description of the model's network architecture, and as a description of the hardware circuitry. The last two options were already discussed and dismissed in previous paragraphs. A functional analysis of the ML system would require appealing to the underlying mechanism, but that is impossible because there is no genuine mechanism to speak of. Therefore, the New Mechanistic view seems inapplicable in the context of DNNs.

Given the obvious objections to the application of a physical or mechanistic theory of causation to DNNs, one wonders why Erasmus et al. (2021) did not adopt a different causal approach, one more amenable to the context at hand. In particular, it would seem more natural to use a counterfactual theory of causal dependence such as Lewis's (1973), and a theory of counterfactual explanation, such as Woodward's (2003). Some recent approaches to XAI have used precisely this approach (Buijsman, 2022; Wachter et al., 2018; Watson & Floridi, 2021; see Durán, forthcoming, for an overview). The reason they offer is that counterfactual approaches to explanation are highly pragmatic and therefore incompatible with the indefeasibility thesis about explanation (Erasmus et al., 2021, p. 839). But if the indefeasibility thesis is false, at least in the context of DNNs, then it ceases to be a reason to reject this approach. The counterfactual approach to explanation in ML has its own problems, which I will discuss in the final section of the chapter, but at least it does not depend on the existence of physical laws, mechanisms, or processes, which seem completely foreign in the context of machine learning.

## 4. A Plea for Understanding

If the attempt to adapt an extant account of scientific explanation to ML is a hopeless endeavor, there are three remaining options: (i) either to adopt a consensual or stipulative definition of "explanation" in ML; (ii) to abandon the factivity condition for explanation; or (iii) to abandon the idea that there is a unique way of understanding what an explanation is in the context of ML. The first option seems entirely unworkable and arbitrary. The second one is mostly associated with pragmatic theories of explanation (Achinstein, 1983; van Fraassen, 1980). These theories have been fruitfully used to clarify the pragmatic context in which explanations are sought in AI (Miller, 2019, 2021). However, there is a tendency to analyze the concept in terms of its empirical usage, without much normative concern.[8] In previous work (Páez, 2019), I have defended the third option. I have argued that XAI ought to take a turn towards a more pragmatic approach in which the focus of attention shifts from the explanation to the understanding of ML systems. If we focus on the cognitive and practical needs of the different stakeholders involved in designing, implementing, and using a ML model, there will be a wide variety of options available to make the model and its outputs understandable. Whether one calls these paths to understanding "explanations" becomes largely irrelevant. In a well-known survey of XAI methods, Guidotti et al. (2018) conclude that one of the most important open problems in XAI is that there is no agreement on what an explanation is:

> Indeed, some works provide as explanation a set of rules, others a decision tree, others a prototype (especially in the context of images). It is evident that the research activity in this field is not providing yet a sufficient level of importance in the study of a general and common formalism for defining an explanation, identifying which are the *properties* that an explanation should guarantee, e.g., soundness, completeness, compactness and comprehensibility (p. 36).

---

[8] I offer a critique of purely pragmatic theories of scientific explanation in Páez (2006).

I am highly skeptical that a formal definition of explanation in terms of necessary and sufficient conditions or properties can be found. Instead of trying to define "explanation" in ML, it would be more fruitful to think of the provision of understanding as the common element that "defines," in a sense, what all XAI methods have in common. This allows a plurality of explanatory methods to flourish as long as they provide understanding to the system's users. A robust account of understanding in ML should provide an adequate grounding for all such methods. In this final section, I want to further develop the account of understanding in ML that I offered in my 2019 paper by adding the idea that understanding is a *success concept* in the sense explained below.

**4.1 Understanding as a Success Concept**

In epistemology, understanding is often distinguished from knowledge.[9] There are two main differences between these concepts. First, understanding is seen as a higher epistemic achievement than knowledge (Kvanvig, 2003; Pritchard, 2010). I can come to know that my coffee is cold just by sipping from the cup, while understanding why my coffee is cold involves relating the coffee's temperature to the laws of thermodynamics. Secondly, the objects of understanding are generally more complex and structured than the objects of knowledge (Zagzebski, 2019). We want to understand the stock market, the theory of relativity, or the New York subway system. Even when we try to understand something simple like my coffee getting cold, the fact must be inserted into a broader, more complex theoretical context.

---

[9] Needless to say, this opinion is not unanimous among philosophers. I cannot discuss all the arguments for and against this view in this chapter, but if my claim that there are non-factive paths to understanding in ML is correct, the absence of truth will prevent these cases of understanding from being reduced to some species of knowledge.

Some philosophers have argued that understanding is simply knowledge of causes: if I know the cause of *p*, I understand why *p* (Lipton, 2004). Similarly, Khalifa (2017) argues that understanding is knowledge of explanations. De Regt offers the following example to show that *knowing* why *p* is not equivalent to *understanding* why *p*:

> Merely knowing that global warming is caused by the increase of $CO_2$ in the atmosphere does not yet amount to understanding it. A student may be able to answer the question "Why does global warming happen"? correctly by answering "Because of the increase of $CO_2$ in the atmosphere". But this does not imply that she understands why global warming occurs —she merely knows what its cause is. … The student understands why global warming happens if she not only knows that it is caused by the increase of $CO_2$, but also "grasps" the causal, explanatory relation between cause and effect. In this case, she needs a theory or model of the climate system that includes the greenhouse effect (2023, pp. 19-20).

Understanding thus requires connecting different pieces of knowledge. Kvanvig explains: "Understanding requires the grasping of explanatory and other coherence-making relationships in a large and comprehensive body of information. One can know many unrelated pieces of information, but understanding is achieved only when informational items are pieced together by the subject in question" (2003, p. 192). In a similar vein, Zagzebski argues that understanding "involves grasping relations of parts to other parts and perhaps the relation of parts to a whole" (2009, p. 144). The kinds of relations she has in mind can be spatial, temporal, or causal. "It seems to me that one's mental representation of the relations one grasps can be mediated by maps, graphs, diagrams, and three-dimensional models in addition to, or even in place of, the acceptance of a series of propositions" (p. 145). For Grimm, understanding a complex object such as the New

York subway system implies an apprehension of how a thing works, "how the various elements of the thing relate to, and depend upon, one another" (2011, p. 86).

Now, what are the conditions of satisfaction of the "grasping" often mentioned in these characterizations of understanding? Traditional opponents of understanding as an interesting philosophical notion, from Hempel (1965) to Trout (2002), often dismissed understanding as a psychological, subjective state that cannot be rigorously analyzed. This attitude is less common today, but there is still no agreement on what the grasping entails. A useful way of thinking about the conditions of satisfaction for grasping the interconnectedness of different facts is in terms of an agent's success in *using* the information. This idea has been defended in different guises by philosophers of science like Ylikoski (2009), De Regt (2017), and Kuorikoski (2011, 2023). In the inferential conception of understanding defended by Kuorikoski and Ylikoski (2015), for example, understanding "is not only about learning and memorizing true propositions, but about the capability to put one's knowledge to use. To understand is to be able to tell what would have happened if things had been different, what would happen if certain things were changed, and what ways there are to bring about a desired change in the outcome" (p. 3819). More specifically, understanding can be equated "with the ability to draw correct counterfactual what-if inferences about the object of understanding. … To understand a phenomenon is to be able to correctly situate it within a space of possibilities" (Kuorikoski, 2023, p. 218). De Regt also argues that genuine understanding manifests itself as a skill: an agent must have the ability to use his knowledge.[10] Among the most important uses of knowledge is the construction of simple, idealized models of some complex phenomenon, which serve as mediators between abstract theories and empirical data. The construction of a successful model with the right idealizing assumptions requires an understanding of the interconnectedness of the data and of the way in which the theory can be applied to the model. Successfully putting one's knowledge to use is not limited to reasoning counterfactually and building models. Being able to fix or improve a system, to profit from it, or to game it, are examples of the many possible ways in which usage is a sign of understanding.[11]

---

[10] But see Sullivan (2018) for an argument against this claim

[11] It is worth noting that Wachter, Mittelstadt and Russell, the proponents of one of the most influential counterfactual methods in XAI, drive a wedge between understanding and use. They

An additional pragmatic aspect of this approach is that there is no unique way of measuring success in using a body of knowledge. There is no unique benchmark for understanding. In De Regt's view, understanding is contextual. Criteria for understanding and intelligibility depend on the historical and disciplinary context:

> In the seventeenth century, for example, the generally accepted view was that only a mechanics based on contact action is intelligible, whereas in the eighteenth century (as a result of the success of Newton's theory of gravitation) action-at-a-distance was regarded as the paradigm of intelligibility ... Of course, one might claim that only one of these positions is the correct one, but that would not do justice to the history of science (2023, p. 19).

Kuorikoski (2023) defends a similar idea: "Historical ruptures in foundational metaphysics, such as the shift from Aristotelian species-essences to Darwinian population thinking or from the classical mechanical conception of the physical reality to the quantum field theory have also constituted fundamental changes in the very criteria of understandability" (pp. 217-218). Furthermore, new sub-fields of science, such as machine learning, "are often built on specific conceptions of scientific understanding" (p. 217). As we will see in the next section, the contextual aspects of understanding in ML cut even deeper.

Despite their acceptance of different conceptions of understandability, Kuorikoski and Ylikoski are fully committed to a factive conception of understanding: drawing incorrect counterfactual inferences would presumably be a sign of a lack of understanding. However, it is difficult to see how a modal approach based on counterfactuals can preserve the factivity condition. Many counterfactual statements are empirically unverifiable and can only be accepted as possible consequences of a scientific theory, or at least of a causal model, which itself can never be absolutely confirmed. De Regt also uses the ability to infer counterfactuals as a criterion for understanding, but correctly rejects the factivity condition. In his approach, theories used to reason counterfactually have to be intelligible and the inferences drawn from them must conform

---

argue that explanations in ML should be "a means to help a data-subject *act* rather than merely understand" (2018, p. 843).

to the basic epistemic values of empirical adequacy and internal consistency. But both intelligibility and empirical adequacy fall short of truth. Naturally, the discussion of whether understanding is factive is not limited to the nature of counterfactual inferences, and the issue has been extensively discussed in epistemology and the philosophy of science. I cannot do justice to the problem here, but this much is clear: an approach to understanding based on pragmatic success need not be subject to the strict alethic standards of knowledge.

Finally, this approach also has the advantage that a person's understanding of a phenomenon or of a subject matter is empirically verifiable. Ylikoski (2009) puts the matter thus:

> When we evaluate somebody's understanding, we are not making guesses about his or her internal representations, but about the person's ability to perform according to set standards. The concept of understanding allows that the ability can be grounded in various alternative ways, as long as the performance is correct. Furthermore, the correctness of the internal model is judged by the external displays of understanding, not the other way around. This makes understanding a behavioral concept (p. 102).

Having intersubjective criteria for a person's understanding has the advantage of providing ways to test the effectiveness of different models, methods, and devices that aid with understanding a phenomenon. Since these criteria are contextual, this approach also allows for the design of tools that aid comprehension with different users and populations in mind (Lage et al., 2019). This will prove to be essential when we think of the different stakeholders involved in the use of machine learning systems.

**4.2 Pragmatic Understanding in ML**

According to the account provided in the previous section, understanding a complex object requires being able to identify its various parts and their interdependence to a degree that enables the agent to use that knowledge in different ways. This is precisely what surrogate models in ML help bring about. Algorithms such as rule lists, sparse decision trees or linear models provide a simplified version of the original model by identifying its main features, the possible interactions between them, and their combined effect on the output. Knowledge of these individual elements without grasping their interconnectedness will not provide understanding of the original model. Furthermore, decision trees are structures that allow counterfactual reasoning. A user can think counterfactually simply by following paths in the tree that do not lead to the original output. Rule lists can also cover counterfactual cases, although this requires a much less intuitive combination of rules.[12] Both designing a surrogate model and using it correctly are evidence that the person understands the target model. A person's understanding of the target model admits of degrees, depending on the person's grasp of the connections displayed in the surrogate model.

Consider an example. Letham et al. (2015) introduced interpretable prediction models that take the form of sparse decision lists. Each list consists of a series of association rules in the form of if-then statements. The rules are pre-mined from the input space using the FP-Growth algorithm (Borgelt, 2005). The resulting models have the same level of complexity as medical scoring systems, thus making them easy to use for clinical practitioners. In terms of performance in a stroke risk classification task, the decision lists were comparable to that of support vector machines (SVMs), and not substantially worse than $L^1$ logistic regression and random forests trained on the same data. Other examples of simple surrogate models are the decision trees for diabetes risk prediction introduced by Bastani et al. (2019) and the checklists proposed by Jung et al. (2017).

Like most models in the social and natural sciences, surrogate models in ML are not factive. They are "true enough" of their target, to use Elgin's (2017) suggestive phrase. Potochnik (2017) argues that in many cases the commonality between representations and what they represent can be understood in terms of functional similarities. Following Potochnik, I think that the relation between surrogate models and their ML targets should be seen in functional terms as well. Functional representation is a pragmatic concept that depends on the specific functions of interest to the modeler or user. Understanding function means being able to build or use the model and manipulate its features to obtain the desired result. We can think of surrogate models as epistemic tools (Currie, 2017; Knuuttila, 2011) designed to capture functional similarities of interest. A correct use of the tool will reveal that the user understands the function of the parts and their overall fit with the target.

---

[12] Lakkaraju et al. (2016) have shown that decision trees in general are easier to understand than rule lists.

In the previous section, we saw that drawing counterfactual inferences is often seen as a sign of understanding. So, why not use local counterfactual XAI methods directly instead of global surrogate models to achieve understanding of the target model? There are several reasons. The first one is that counterfactual methods suffer from a lack of robustness. Like perturbation-based methods, local counterfactual methods can be easily manipulated and may converge toward drastically different explanations under small perturbations (Slack et al., 2021). Counterfactual probes also critically depend on closeness metrics but there is no principled way to decide which metric to use in any given case, especially since most variables are not causally independent. And like saliency-based methods, the lack of causal grounding can deliver sub-optimal or even erroneous explanations to decision-makers (Chou et al., 2022; Karimi et al., 2022).

To complicate things, there seems to be a lack of interest in the AI community to test whether the counterfactual methods they employ are actually understood by non-experts. A recent survey shows that only 36 out of 117 (31%) research papers evaluating counterfactual explanations included user studies (Keane et al., 2021). As the authors state in the beginning of the paper, "the XAI community is busily developing technical solutions that may have no practical benefits to people in real-life" (p. 1).

The third reason is epistemologically deeper. Genuine counterfactuals are theory or model dependent. Counterfactuals need a possibility space that can only be defined using lawlike statements and background conditions, or by means of a fully specified causal model. Otherwise, there will be no basis to answer what-if-things-had-been-different questions. Furthermore, as Beckers (2022) points out, understanding the space of possibilities also requires understanding which factors cannot be changed or manipulated in a counterfactual situation. Current counterfactual methods in XAI make simplifying assumptions that land them far from the required conditions to qualify as based on bona fide counterfactuals. In particular, many methods assume that the input variables are causally independent. The strategy is to keep all variables at a fixed value except one and see how a change in that variable produces a different output.[13] Wachter et al. (2018), for example, define an explanation of the output of an algorithm as having the following format: "Score $p$ was returned because variables $V$ had values ($v_1$, $v_2$, ...) associated with them. If $V$ instead had values ($v_1'$, $v_2'$, ...), and all other variables had remained constant, score $p$ would have been returned" (p. 848). The method tells you how to change the output, but not why the output changes. It does not provide an understanding of the process because there is no grasp of the logical and causal connections between the features, and between the features and the prediction. Again, *knowing* why $p$ changed is not the same as *understanding* why $p$ changed.

---

[13] Karimi et al. (2022) suggest that "it is perhaps more appropriate to refer to these approaches as contrastive, rather than counterfactual explanations" (p. 360).

A similar argument can be used against local saliency and perturbation methods: they tell users which features are relevant to an individual prediction but not why they are relevant, that is, users *know* what caused the output, but they do not *understand* why. The lack of understanding is demonstrated by the difficulties users have in predicting the "explanation" for new, similar input-output pairs. In an empirical study of saliency maps generated by a layer-wise relevance propagation (LRP) algorithm (Bach et al., 2015) on an image classification task using a convolutional neural network, Alqaraawi et al. (2020) found that "the maps seem to provide very limited help for participants to anticipate the network's output for new images" (p. 275). Post hoc local interpretability methods are useful for developers trying to adjust the parameters of a model, but they seem of limited use for other stakeholders.

What is missing from all of these local methods is a grasp of the causal and logical relationships in the input space. Users must build a *mental model* of the relevant features and the possible outcomes before they can make inferences about the target model. In an experiment designed to measure the effectiveness of different global XAI methods, van der Waa et al. (2021) used two behavioral measures of a participant's understanding of the model:

> The first behavioral measurement assessed the participant's capacity to correctly identify the decisive factor of the situations in the system's advice. This measured to what extent the participant recalled what factor the system believed to be important for a specific advice and situation. Second, we measured the participant's ability to accurately predict the advice in novel situations. This tested whether the participant obtained a mental model of the system that was sufficiently accurate enough to predict its behavior in novel situations (pp. 16-17).

The authors found that presenting participants only with counterfactuals about the system's advice did not improve their identification or predictive abilities when compared to no explanations at all. When presented with rule-based explanations without counterfactuals, the first behavioral measure of understanding improved, but not the second. Lim et al. (2009) also present empirical evidence that suggests that offering users of ML systems decision rules encoded in decision trees yields superior results in terms of both understanding and trust when compared to counterfactual explanations. Buijsman (2022) uses this evidence to conclude that "a counterfactual is only helpful when it suggests a reasonable generalization. … A pure case-by-case approach, where counterfactuals are presented but without overarching generalizations, doesn't truly explain the functioning of an algorithm" (p. 569).

In the context of algorithmic recourse, Karimi et al. (2022) also argue that the possibility of changing an unfavorable outcome requires causal knowledge. The authors attribute the shortcomings of existing XAI counterfactual approaches "to their lack of consideration for real-world properties, specifically the causal relationships governing the world in which actions will be performed" (p. 354). The question is, in what form should causal relations be displayed to be understandable to lay users, and how precise and comprehensive should those causal relationships be? It is well-known that is extremely difficult to specify a full causal model. Despite requiring real-world causal relationships, Karimi et al. (2022) admit that "in practice, the underlying causal model is rarely known" (p. 360). Since it is extremely difficult to build a factual causal model, algorithmic recourse also needs the understanding provided by non-factual, simpler surrogate models.[14]

As I see it, a surrogate model needs to be "causal enough"—to paraphrase Elgin's phrase— for a user to understand how the parts are interconnected and interdependent, but context will determine the depth of the causal knowledge that needs to be displayed in the model. In scientific contexts, one should expect a highly detailed map of the causal relationships, based on solid background scientific knowledge (Sullivan, 2022), but other contexts might only require sparse decision trees with very few variables and a higher tolerance for spurious correlations. The complexity of the model will be determined by the explanatory requirements of the stakeholders involved (Zednik, 2021). Even the normative standards for transparency that have been proposed recently, such as IEEE P7001 (Winfield et al., 2021), recognize that transparency depends entirely on the background knowledge of the stakeholders trying to understand the model. Specifying the exact shape that these causal surrogate models should take is a very difficult challenge that falls beyond the scope of this chapter, but causal machine learning is a very active field of research that might provide better paths toward understandable AI.

---

[14] Sullivan and Kasirzadeh (chapter 8 in this volume) go even further and argue that providing users with explanations that aim at understanding AI decisions is (ethically) preferable than providing them with recourse explanations.

## 5. Conclusion

The use of the term "explanation" in AI is a conceptual jumble. I believe no amount of conceptual analysis will result in a clear definition or in a more widely accepted set of usage rules. In this chapter, I have shown that the attempt to ground the term in the way it is used in the philosophy of science is unsuccessful and wrongheaded. Explanations in AI are neither logical nor physical nor mechanical nor nomological. They are not even factual. A much better option is to abandon the definitional enterprise and focus instead on the result of what "explanations" are supposed to achieve: understanding of opaque machine learning systems. I have argued for a pragmatic notion of understanding that is empirically testable and broad enough to allow many different approaches and methods.

## References

Achinstein, Peter. 1983. *The nature of explanation*. New York: Oxford University Press.

Alqaraawi, Ahmed, Martin Schuessler, Philipp Weiß, Enrico Costanza, and Nadia Berthouze. 2020. Evaluating saliency map explanations for convolutional neural networks: a user study. In *IUI '20: Proceedings of the 25th International Conference on Intelligent User Interfaces*, 275-285. New York: ACM.

Ancona, Marco, Enea Ceolini, Cengiz Öztireli, and Markus Gross. 2019. Gradient-based attribution methods. In *Explainable AI: Interpreting, explaining and visualizing deep learning*, ed. W. Samek, G. Montavon, A. Vedaldi, L. K. Hansen, and K.-R. Müller, 169-191. Cham: Springer.

**Acknowledgements**

I would like to thank Emily Sullivan and Juan M. Durán for their careful reading of the manuscript and for their detailed comments, which have been immensely useful. I also profited from useful discussions with Victoria Martín del Campo and Stefan Buijsman during the session organized by Juan as part of the XVIIth Congress on Logic, Methodology, and Philosophy of Science and Technology in Buenos Aires. As always, I am indebted to Arnold Koslow, from whom I learned everything I know about scientific explanation.

**Andrés Páez** is Professor of Philosophy and Researcher in the Center for Research and Formation in Artificial Intelligence (CinfonIA) at the Universidad de los Andes in Bogotá, Colombia. His current areas of research are the philosophy of artificial intelligence, legal and social epistemology, and the philosophy of science. He is the former President of the Latin American Association for Analytic Philosophy (ALFAn) and member of the editorial board of *Minds & Machines*.